\documentclass[11pt]{article} % For LaTeX2e
\usepackage{rldmsubmit,palatino}
\usepackage{algorithm}
\usepackage{algorithmic}
\usepackage{amsmath}
\usepackage{amsfonts}
\usepackage{dsfont} 
\usepackage{hyperref}
\usepackage{url}
\usepackage{graphicx}
\usepackage{subcaption}

\usepackage[colorinlistoftodos]{todonotes}
\definecolor{flame}{rgb}{0.89, 0.35, 0.13}

\usepackage[round]{natbib}

\title{Actor-Critic with Active Importance Sampling}

\author{
Majid Molaei \\
Politecnico di Milano\\
Milan, Italy \\
\texttt{majid.molaei@polimi.it} \\
\And
Gabor Paczolay \\
Politecnico di Milano \\
Milan, Italy \\
\texttt{paczolay.gabor@gmail.com} \\
\AND
Matteo Papini \\
Politecnico di Milano \\
Milan, Italy \\
\texttt{matteo.papini@polimi.it} \\
\And
Alberto Maria Metelli \\
Politecnico di Milano \\
Milan, Italy \\
\texttt{albertomaria.metelli@polimi.it} \\
\And
Marcello Restelli \\
Politecnico di Milano \\
Milan, Italy \\
\texttt{marcello.restelli@polimi.it} \\
}

\begin{document}

\maketitle

\begin{abstract}
 This paper introduces the Active-Importance-Sampling Actor-Critic (AISAC) algorithm, an extension of the Actor-Critic framework in reinforcement learning to reduce variance in policy gradient estimation. Actor-Critic methods combine policy optimization (actor) with value function estimation (critic), using neural networks for both. However, traditional methods often suffer from high gradient variance, limiting efficiency and stability.

AISAC mitigates this by optimizing the behavior policy to minimize gradient variance. Leveraging importance sampling principles from Monte Carlo theory, AISAC selects a behavior policy approximating the ideal data-collecting distribution. Aligning the behavior policy with the gradient magnitude and direction of the target policy ensures efficient data collection and faster convergence.

We provide a theoretical analysis showing AISAC reduces variance while maintaining unbiased policy gradient estimation. For continuous action spaces, AISAC uses Gaussian distributions optimized via cross-entropy minimization.

Experiments on Inverted Pendulum and Half Cheetah tasks demonstrate AISAC’s effectiveness, achieving faster learning, better sample efficiency, and greater stability than standard Actor-Critic algorithms. The optimized behavior policy improves both the target policy and critic accuracy, delivering superior performance across hyperparameter settings.

AISAC stabilizes training and accelerates convergence, making it suitable for real-world reinforcement learning applications. Rigorous simulations and statistical analysis confirm the reliability of these gains.

In conclusion, AISAC offers a robust solution for reducing variance in policy gradient estimation, enhancing reinforcement learning’s efficiency and stability. Future work will explore its integration with advanced algorithms like Soft Actor-Critic and Twin Delayed Deep Deterministic Policy Gradient to address complex, high-dimensional tasks.
\end{abstract}


\startmain % to start the main 1-4 pages of the submission.

\section{Introduction}
Actor-Critic (AC) algorithms~\citep{konda1999actor} combine the flexibility of policy gradient methods~\citep{agarwal2021theory} with the representational power of value-based approaches~\citep{mnih2015human}, achieving success in tasks ranging from continuous control to discrete environments~\citep{haarnoja2018soft,lillicrap2016continuous,fujimoto2018addressing}. AC agents consist of an actor, which selects actions, and a critic, which evaluates them to guide policy updates. Off-policy learning~\citep{lillicrap2016continuous} decouples the data collection policy (behavior policy) from the optimized policy (target policy), enabling variance reduction and improved performance. Building on recent work~\citep{papini2024policy}, we extend the active importance sampling (AIS) technique~\citep{hanna2017data,metelli2023relation} to AC algorithms, optimizing the behavior policy to collect high-quality data for variance-reduced gradient estimation.

We propose the Active-Importance-Sampling Actor-Critic (AISAC) algorithm, which aligns the behavior policy with an ideal data-collecting distribution. Theoretical analysis demonstrates the variance-reduction properties of AISAC, and experiments on continuous control tasks confirm its advantages in sample efficiency and learning performance. The paper is organized as follows:
Section~\ref{sec:idea} presents the theoretical study, Section~\ref{sec:algo} introduces the AISAC algorithm, and Section~\ref{sec:exp} discusses experimental results.

\section{Preliminaries}\label{sec:pre}

The \textbf{Policy Gradient Theorem}~\citep{sutton1999policy} provides a foundation for optimizing policies in reinforcement learning. The average reward \( r(\pi) \) under a policy \(\pi\) is expressed as:

\begin{equation}\label{1}
 r(\pi) = \sum_{s} \mu(s) \sum_{a} \pi(a \mid s, \theta) \sum_{s', r} p(s', r \mid s, a) r,   
\end{equation}

where \(\mu(s)\) is the initial state distribution, \(\pi(a \mid s, \theta)\) is the policy parameterized by \(\theta\), and \(p(s', r \mid s, a)\) represents the transition dynamics. To optimize the policy, the gradient of \( r(\pi) \) with respect to \(\theta\) is computed as:

\begin{equation}\label{2}
  \nabla_{\theta} r(\pi) = \mathbb{E}_{\pi} \left[ \nabla_{\theta} \log \pi(a \mid s, \theta) Q^{\pi}(s, a) \right], 
\end{equation}

where \( Q^{\pi}(s, a) \) is the action-value function.

The \textbf{Actor-Critic Algorithm}~\citep{konda1999actor} combines policy optimization (actor) and value function estimation (critic). The actor updates policy parameters \(\theta\) to maximize rewards, while the critic refines the value function. The algorithm iteratively selects actions using \(\pi(a \mid s, \theta)\), and computes the Temporal Difference (TD) error:

\[
\delta = R + \gamma \sum_{a'} \pi(a' \mid S', \theta) Q^{\pi}(S', a', w) - Q^{\pi}(S, A, w),
\]

and updates the critic and actor parameters as:

\[
w \leftarrow w + \alpha^w \delta \nabla_w Q^{\pi}(S, A, w),
\quad
\theta \leftarrow \theta + \alpha^\theta \nabla_{\theta} \ln \pi(A \mid S, \theta) Q^{\pi}(S, A, w).
\]

\textbf{Importance Sampling} addresses inefficiencies in Monte Carlo integration by using an alternative distribution \( q(x) \) to estimate expectations under \( p(x) \). The estimator:

\[
\hat{\mu}_q = \frac{1}{n} \sum_{i=1}^n \frac{f(X_i)p(X_i)}{q(X_i)}, \quad X_i \sim q,
\]

is unbiased, with variance dependent on \( q(x) \). Optimal \( q(x) \) minimizes variance by being proportional to \( |f(x)|p(x) \), though practical implementation may require approximations~\citep{mcbook}.

\section{Idea}\label{sec:idea}

By considering the expected reward \( r(\pi) \) under a policy \(\pi\) in Equation~\eqref{1}, we seek a behavior policy that reduces the sample complexity in estimating the gradient of the average reward objective in Equations~\eqref{2}.

The integral to estimate is:
\[
I = \mathbb{E}_{\pi(a \mid s, \theta)}\left[ \nabla \ln \pi(a \mid s, \theta) Q^{\pi}(s, a) \right].
\]

Rewriting using an importance sampling distribution \( b(a \mid s) \):
\[
I = \mathbb{E}_{b(a \mid s)}\left[ \frac{\nabla \ln \pi(a \mid s, \theta) Q^{\pi}(s, a)}{b(a \mid s)} \right].
\]

The estimator becomes:
\[
\hat{I}_{\text{IS}} = \frac{1}{n} \sum_{i=1}^{n} \frac{\nabla \ln \pi(a_i \mid s, \theta) Q^{\pi}(s, a_i)}{b(a_i \mid s)},
\]
where \( a_i \sim b(a \mid s) \).

\subsection*{Variance of the Estimator}

The variance of \( \hat{I}_{\text{IS}} \) is:
\[
\mathrm{Var}(\hat{I}_{\text{IS}}) = \frac{1}{n} \left( \int \frac{(\nabla \pi(a \mid s, \theta) Q^{\pi}(s, a))^2}{b(a \mid s)} \, da - I^2 \right).
\]

\subsection*{Reduction in Variance}

To compare, the Monte Carlo variance is:
\[
\mathrm{Var}(\hat{I}) = \frac{1}{n} \left( \int (\nabla \pi(a \mid s, \theta) Q^{\pi}(s, a))^2 \pi(a \mid s, \theta) \, da - I^2 \right).
\]

Importance sampling reduces variance if:
\[
\int \frac{(\nabla \pi(a \mid s, \theta) Q^{\pi}(s, a))^2}{b(a \mid s)} \, da < \int (\nabla \pi(a \mid s, \theta) Q^{\pi}(s, a))^2 \pi(a \mid s, \theta) \, da.
\]

As shown in~\citep{mcbook}, \( b(a \mid s) \) should be chosen such that \( \frac{\pi(a \mid s, \theta)}{b(a \mid s)} \) is minimized where \( \nabla \ln \pi(a \mid s, \theta) Q^{\pi}(s, a) \pi(a \mid s, \theta) \) is large. This ensures \( b(a \mid s) \) approximates \( \pi(a \mid s, \theta) \) weighted by \( \nabla \ln \pi(a \mid s, \theta) Q^{\pi}(s, a) \).

However, for policy gradient vectors, we must consider both magnitude and direction. The behavior policy should be designed so that, in state \( s \), actions \( a \) aligning the target policy gradient direction with the expected vector direction are favored. This generalizes active importance sampling from scalar to vector functions, optimizing sample efficiency and variance reduction.

Thus, by incorporating the dot product of these two vectors into the previous definition, the behavior policy becomes proportional to:

\begin{align}
 &b(a\mid s)\propto \Big|[\nabla_\theta \ln \pi(a\mid s, \theta)Q^{\pi}(s, a)] \cdot [\sum\limits_{a}\nabla_\theta \pi(a\mid s, \theta)Q^{\pi}(s, a)]\Big| \times \pi(a\mid s, \theta)(\mathds{1}_{\{\pi(a\mid s, \theta) \neq 0\}}) \quad \quad \quad \quad \forall (s,a)
 \nonumber\\[8pt]
 &\Rightarrow b(a\mid s)\propto \Big|\big([\nabla_\theta \pi(a\mid s, \theta)]\cdot [\sum\limits_{A}\nabla_\theta \pi(A\mid s, \theta)Q^{\pi}(s, A)]\big) \times Q^{\pi}(s, a)\Big|(\mathds{1}_{\{\pi(a\mid s, \theta) \neq 0\}}) \ \quad \quad \quad \quad \quad \quad \forall (s,a).
  \nonumber
\end{align}

Once again, for every state-action pair, we can normalize this result by dividing it by the sum of all possible actions in a specific state; thus we reach the behavior policy (in tabular form):

\begin{align}
&b(a \mid s)=\frac{\Big|\big([\nabla_{\theta} \pi(a \mid s, \theta)] \cdot [\sum\limits_{A} \nabla_{\theta} \pi(A \mid s, \theta) Q^{\pi}(s, A)]\big) \times Q^{\pi}(s, a)\Big|\mathds{1}_{\{\pi(a \mid s, \theta) \neq 0\}}}{\sum\limits_{A} \bigg( \Big| \big([\nabla_{\theta} \pi(A \mid s, \theta)] \cdot [\sum\limits_{A} \nabla_{\theta} \pi(A \mid s, \theta) Q^{\pi}(s, A)]\big) \times Q^{\pi}(s, A)\Big| \mathds{1}_{\{\pi(a \mid s, \theta) \neq 0\}}\bigg)} \nonumber
\end{align}

At this point, we have achieved the goal of finding the behavior policy that reduces the number of samples needed to compute the gradient estimation of our average reward objective.

At this stage, we can proceed to implement our proposed algorithm to observe the results in practice.

\section{Algorithm}\label{sec:algo}

Algorithm~\ref{alg:actor-critic-with-AIS} uses a parameterized policy to maximize expected returns by refining the action-value function, which estimates the expected return of actions in specific states. In each iteration, a behavior policy promotes exploration, and the agent samples an action based on this policy, observing the resulting state and reward. The temporal difference error updates the action-value function, while policy parameters are adjusted using off-policy gradients. For continuous actions, we use a Gaussian policy with mean $\mu(S, \theta)$, and the behavior policy is proportional to:

\[
\Big| \big( [\nabla_{\theta} \pi(a \mid s, \theta)] \cdot [\int_{a} \nabla_{\theta} \pi(a \mid s, \theta) Q^{\pi}(s, a, w) \, da] \big) \times Q^{\pi}(s, a, w) \Big|.
\]

This simplifies, for Gaussian policies, to:

\[
\Big| \big( [\nabla_{\theta} \pi(a \mid s, \theta)] \cdot [\nabla \mu(s, \theta) \nabla_a Q^\pi(s, a, w)] \big) \times Q^\pi(s, a, w) \Big|.
\]

This expression allows for the use of cross-entropy minimization, eliminating the need for explicit normalization and integration~\citep{papini2024policy}.

\begin{algorithm}[t]
\caption{Actor-Critic Algorithm with Active IS}
\label{alg:actor-critic-with-AIS}
\begin{algorithmic}[1]

\STATE \textbf{Input:} a differentiable policy parameterization, e.g. Gaussian \( \pi(a \mid s, \theta) = \frac{1}{\sqrt{2\pi\sigma_\theta^2(s)}} \exp\left(-\frac{(a - \mu_\theta(s))^2}{2\sigma_\theta^2(s)}\right) \) 
\STATE \textbf{Input:} a differentiable action-value function parameterization \( q : S \times A \times \mathbb{R}^{d} \to \mathbb{R} \)
\STATE \textbf{Initialize} value-function weights \( w \in \mathbb{R}^{d} \) and policy parameter \( \theta \in \mathbb{R}^{d'} \) arbitrarily
\STATE \textbf{Algorithm parameters:} \( \alpha^{w} > 0 \),  \( \alpha^{\theta} > 0 \)
\STATE \textbf{Initialize} \( s \in S \)
\STATE \textbf{Loop forever and for each time step:}
\STATE 
\begin{align*}
&b(a \mid s)\propto\nonumber %\\[8pt]
%&
\quad \left|\left(\nabla_{\theta} \pi(a \mid s, \theta) \cdot \int_{a} \nabla_{\theta} \pi(a \mid s, \theta) Q^{\pi}(s, a, w)\bigg. \, da\right)  Q^{\pi}(s, a, w)\right|\mathds{1}_{\{\pi(a \mid s, \theta) \neq 0\}}%\\
%&\frac{
%\bigg. \times Q^{\pi}(s, a, w)\Big|\mathds{1}_{\{\pi(a \mid s, \theta) \neq 0\}}
%}{\int_{a} \bigg( \Big| \big([\nabla_{\theta} \pi(a \mid s, \theta)] \cdot [\int_{a} \nabla_{\theta} \pi(a \mid s, \theta) Q^{\pi}(s, a, w) \, da]\big) \bigg.} \nonumber \\
%&\quad \quad \quad \quad \bigg. \times Q^{\pi}(s, a, w)\Big| \mathds{1}_{\{\pi(a \mid s, \theta) \neq 0\}}\bigg) \, da \nonumber
\end{align*}

\STATE \quad Take action $A$:\\
\quad \quad $A \sim b(a \mid s)$
\STATE \quad Observe $S', R$:

\STATE \quad \quad $\delta \leftarrow R + \gamma \int_{a'} \pi(a' \mid s', \theta) Q^{\pi}(s', a', w) \, da'-Q^{\pi}(s, A, w)$
\STATE \quad \quad $w \leftarrow w + \alpha^{w} \delta \nabla_w Q^{\pi}(s, a, w)$

\STATE \quad \quad $\theta \leftarrow \theta + \alpha^{\theta} \frac{\mu_{\pi}(s)\pi(a \mid s, \theta)}{\mu_{b}(s)b(a \mid s)} \nabla_\theta \ln \pi(a\mid s, \theta) \times Q^{\pi}(s, a, w)$

\STATE \quad %\text{Where:}
%\STATE \quad \quad $\nabla_\theta \ln \pi(a \mid S, \theta) = \frac{a - \mu(S, \theta)}{\sigma^2(S, \theta)} \nabla \mu(S, \theta)$\\
%\quad \quad \quad \quad \quad \quad \quad \quad $+ \left( \frac{(a - \mu(S, \theta))^2}{\sigma^3(S, \theta)} - \frac{1}{\sigma(S, \theta)} \right) \nabla \sigma(S, \theta)$

\STATE \quad \quad $s \leftarrow s'$
\end{algorithmic}
\end{algorithm}

\section{Experiments}\label{sec:exp}
 
We conducted numerical experiments on two Mujoco control tasks: Inverted Pendulum and Half Cheetah, comparing an optimized behavior policy with a simple on-policy actor-critic baseline. Inverted Pendulum involves balancing a pole on a sliding base, while Half Cheetah teaches a two-legged robot to run forward. Both tasks are continuous, using parametrized behavior policies and Gaussian distributions. The experiments involved 2500 iterations with 2000 steps per iteration, testing both on-policy and off-policy critics. The off-policy algorithm, labeled AISAC, showed improved results, particularly when using an optimized behavior policy, and was evaluated on 96 cores with 256GB of RAM.

\begin{figure}[t]
    \centering
    \begin{minipage}{0.45\linewidth}  
        \centering
        \includegraphics[width=\linewidth]{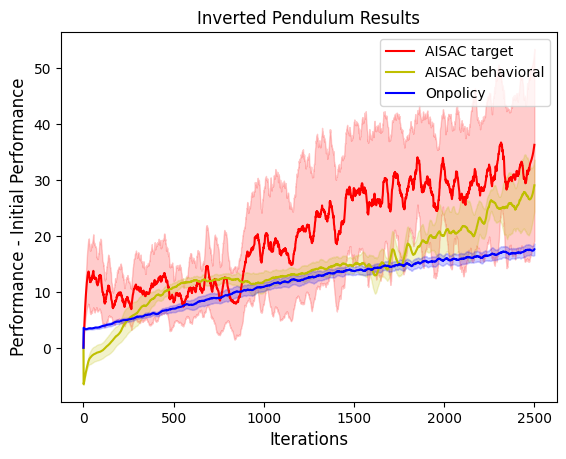}
        \caption{Results for the Inverted Pendulum environment.}
        \label{fig:invpend}
    \end{minipage} \hfill  
    \begin{minipage}{0.45\linewidth}
        \centering
        \includegraphics[width=\linewidth]{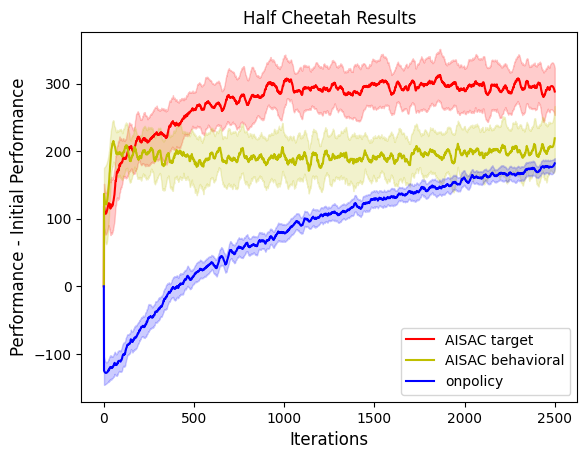}
        \caption{Results for the Half Cheetah environment.}
        \label{fig:hc}
    \end{minipage}
\end{figure}

Figures \ref{fig:invpend} and \ref{fig:hc} show results for the Inverted Pendulum and Half Cheetah environments, with learning curves smoothed using a Savitzky-Golay filter. AISAC outperformed the on-policy actor-critic for both the target \emph{and the behavior policy}, which aligns with \citet{metelli2023relation}'s findings.

\section{Conclusion}

In this paper, we introduced Active-Importance-Sampling Actor-Critic (AISAC), a novel approach applying active importance sampling to the Actor-Critic framework for variance reduction in policy gradient estimation. AISAC improves data collection with an optimized behavior policy, enhancing learning efficiency and accuracy. Our experiments confirm that incorporating variance-reduction techniques accelerates learning in Actor-Critic methods. Future work could extend this idea to advanced actor-critic algorithms like SAC~\citep{haarnoja2018soft} or TD3~\citep{fujimoto2018addressing}.

\bibliographystyle{unsrtnat}
\bibliography{references}

\end{document}